\documentclass[journal]{IEEEtran}
\usepackage{blindtext}
\usepackage{graphicx}
\usepackage{color}
\usepackage{caption}
\usepackage{subcaption}
\usepackage{amssymb}
\usepackage{amsthm}
\usepackage{float}
\usepackage{tikz}

\usepackage{comment}

\newtheorem{thm}{Theorem}[section]

\theoremstyle{definition}
\theoremstyle{remark}

\newtheorem{definition}{Definition}

\ifCLASSINFOpdf
\else
\fi

\usepackage{amsmath}

\usepackage{color}

\begin{document}

\title{RGB image-based data analysis via discrete \\ 
Morse theory and persistent homology}

%
%
%

\author{Chuan~Du*, Christopher~Szul,* \thanks{*Co-first Authors} Adarsh~Manawa, Nima~Rasekh, Rosemary~Guzman, and~Ruth~Davidson** \thanks{**Corresponding Author: redavid2@illinois.edu}}
 


%
%

\markboth{RGB image-based data analysis}%
{}
%


\maketitle

\begin{abstract}
Understanding and comparing images for the purposes of data analysis is currently a very computationally demanding task. A group at Australian National University (ANU) recently developed open-source code that can detect fundamental topological features of a grayscale image in a computationally feasible manner.  This is made possible by the fact that computers store grayscale images as cubical cellular complexes.  These complexes can be studied using the techniques of discrete Morse theory.
We expand the functionality of the ANU code by introducing methods and software for analyzing images encoded in red, green, and blue (RGB), because this image encoding is very popular for publicly available data.  Our methods allow the extraction of key topological information from RGB images via informative persistence diagrams by introducing novel methods for transforming RGB-to-grayscale. This paradigm allows us to perform data analysis directly on RGB images representing water scarcity variability as well as crime variability. We introduce software enabling a a user to predict future image properties, towards the eventual aim of more rapid image-based data behavior prediction. 
\end{abstract}


\thanks{**Corresponding Author}
\thanks{*Co-first authors}

\begin{IEEEkeywords}
Discrete Morse Theory, Image Analysis, Data Behavior Prediction, RGB-to-Grayscale Image Conversion
\end{IEEEkeywords}

\section{Introduction}\label{Introduction}


    
Persistent homology (PH) has been identified as one of the most promising mathematical tools currently available for data analysis \cite{edelsbrunner2008persistent}. For example, PH has been employed for deriving effective methods in machine learning,  \cite{reininghaus2015stable}, and analysis of percolating surfaces and porous materials \cite{robins2016percolating}. Further, PH has been shown to be an effective method for analyzing datasets related to medicine such as the homology of brain networks \cite{lee2012persistent}, brain artery tree structure \cite{bendich2016persistent}, and orthodontics \cite{gamble2010exploring}.  
    
Discrete Morse theory (DMT) has emerged as a tool that can be used in combination with persistent homology for data analysis \cite{chung2009persistence, mischaikow2013morse} because it has been demonstrated that (1) the Betti numbers produced in multi-dimensional persistent homology calculations are \emph{stable} functions, meaning that small perturbations in the data set does not change the  resulting Betti numbers \cite{cerri2013betti} (see Section \ref{sec:PH}, and that (2) discrete Morse theory, as it can identify the Betti numbers of cell complexes, can be used to make persistent homology computations more efficient \cite{mischaikow2013morse, chung2009persistence}.

The novel contribution of this manuscript is the re-purposing of open-source code developed in the publications \cite{robins2011theory, delgado2015skeletonization} (hereafter referred to as the ANU code in this manuscript) that uses a combination of DMT for the \emph{partitioning} and \emph{skeletonizing} (in other words finding the underlying topological features of) images using differences in grayscale values as the distance function for DMT and PH calculations.

The data analysis methods we have developed for analyzing spatial properties of the images could lead to image-based predictive data analysis that bypass the computational requirements of machine learning, as well as lead to novel targets for which key statistical properties of images should be under study to boost traditional methods of predictive data analysis.  
     
The methods in \cite{robins2011theory, delgado2015skeletonization} and the code released with them were developed to handle grayscale images. But publicly available image-based data is usually presented in heat-map data using Red-Green-Blue (RGB) encoding. Since RGB encoding varies by image and uses unequal weighting depending on the image under study, using an ``off the shelf" RGB-to-grayscale conversion (for example, using Preview for grayscale conversion on a Macintosh) before running the DMT+PH analysis will not reliably result in an informative persistence diagram suitable for robust data analysis. We have developed an additional code repository that expands the functionality of the repository released with \cite{robins2011theory, delgado2015skeletonization} to allow for multiple ways (see Section \ref{sec:results} for details) to deal with obstacles to (1) robust data analysis and (2) predictions of data behavior based on image analysis alone.  
     
In particular the methods we have developed with the ANU code allow for multiple paths to enable user-defined variability for conversion of RGB-to-grayscale. This empowers the data-informed user of our methods to solve the problem of variation of RGB encoding of publicly available image data.  This is important as such variation can lead (see Section \ref{sec:scarcity}) to insufficiently informative data analysis. Further, we exploit the speed and stability of their algorithm while allowing wide application to public image data via user-defined inputs for our code as well as the prediction of data behavior. 

For example, such predictive capability has the potential to provide a user-informed image-based alternative to to machine learning methods for weather and climate assessment \cite{priestley1972assessment, xingjian2015convolutional} as well as computationally and statistically costly methods in the social sciences \cite{gill2014bayesian}. We note that computational methods for application-specific data analysis are needed experts in myriad fields. The motive of this publication is to place better computational methods for complex data in the hands of field-specific (in such fields as climate science and many social sciences) experts that can interpret meaningful features of images. 
     
As case studies, we use our methods to analyze patterns in two applications. The first is a proof of concept that user-controlled RGB-to-grayscale conversion affects the tractability of image based data analysis using water scarcity (see Section \ref{sec:scarcity}). Our techniques allow publicly available heat map data from regions with highly variable water scarcity, a common problem for resource management \cite{oki2006global} to be used as an analytical tool for such scarcity. We chose this application because it is an important resource management issue \cite{smakhtin2004pilot} where water resource experts could benefit from image-based methods for policy management, as data-gathering in this field is very difficult and time-consuming \cite{petit2016paradise}. 
     
The second application of our novel RGB-to-grayscale conversion techniques is to heat-map based crime data in Halifax, Nova Scotia (see Section \ref{sec:crimedata}).  We identified this as a potential application because of the abundance of interest in the literature of the statistical properties of data about crime variation \cite{nakaya2010visualising}, as well as interest in fast machine learning techniques (for example) to analyze such data \cite{de2017automatic}. 

\section{Mathematical Foundations}\label{sec:math}

In this section we review the relevant mathematical definitions for our data analysis techniques and computational methods. In particular, we review the relevant defininitions underlying {\it discrete Morse theory} (DMT), the discrete vector fields (DVF) that DMT produces, and how these objects aid in the computation of {\it persistent homology} (PH).

\subsection{Discrete Morse Theory}\label{sec:DMT}

We refer the reader to \cite{forman2002user} for a comprehensive guide to the historical development and concepts of DMT, and largely follow the notation in \cite{forman2002user} for the following definitions.

\begin{definition}\label{def:cellcomplex}
A finite \emph{cell complex} $(V,K)$ is a collection of vertices $V$ where $K$ is a collection of subsets of $V$ satisfying 
\begin{enumerate}
    \item $\{v \in K \ | \text{for all} \ v \in V \}$, and 
    \item if $\alpha$ is a cell defined by subset of the vertices of another cell $\beta \in K$, $\alpha \in K$. 
\end{enumerate}
\end{definition}

We use the notation $\alpha^{p}$ to indicate that the the dimension of the cell $\alpha \in K$ is $p$.

\begin{definition}\label{def:DMF}

A function $f : K \rightarrow \mathbb{R}$ is a \emph{discrete Morse function} (DMF) if for every $\alpha^{(p)} \in K$: 

\vspace{0.2cm}

(1) $|\{\beta ^ {(p+1)} \supsetneq \alpha \mid f (\beta)\leq f (\alpha)\}| \leq 1,$ and

\vspace{0.2cm}

(2) $|\{\gamma ^ {(p-1)} \subsetneq \alpha \mid f (\gamma)\geq f (\alpha)\}|\leq 1$. 

\end{definition}

\vspace{0.2cm}

The use of a DMF allows for the construction of a discrete gradient vector field (a DVF) on a cell complex, which simplifies the computation of the Betti numbers of $K$.  To explain the construction of a DVF we must introduce the notion of critical cells:

\begin{definition}\label{def:critical}

(1) $|\{\beta ^ {(p+1)} \supsetneq \alpha \mid f (\beta)\leq f (\alpha)\}| = 0$, and

\vspace{0.2cm}

(2) $|\{\gamma ^ {(p-1)} \subsetneq \alpha \mid f (\gamma)\geq f (\alpha)\}| = 0$ 
\end{definition}

Observe that non-critical cells come in pairs, which is relevant to the next definition:

\begin{definition}\label{def:DVF}
A \emph{discrete vector field} (DVF) on a cell complex induced by a DMF is defined by arrows pointing from a higher-dimensional cell $\beta^{p}$ to a lower dimensional cell $\alpha^{p-1}$ such that $\alpha^{p-1}$ is \emph{not} assigned a higher value than $\beta^{p}$ by the DMF.  
\end{definition}

The reason we want to find and understand the critical cells follows \cite{robins2011theory}. We refer the reader to \cite{milnor1959spaces} for an introduction to the topological relevance of \emph{CW} complexes. Informally, CW complexes are complexes where cells of different dimension are glued together under sufficiently flexible topological rules. 

\begin{thm}\cite[Theorem 2.5]{robins2011theory}
 Suppose $K$ is a simplicial complex with a DMF.
 Then $K$ is homotopy equivalent to a CW complex  with exactly one cell of dimension $p$
 for each critical simplex of dimension $p$.
\end{thm}

Thus by finding critical cells we are able to recover all the essential topological data of our cell complex. More importantly, by the theorem above, the topological data does not depend on any particular choice of a DMF.  Thus our methods are well-defined after the user enables an RGB-to-grayscale encoding. 

\begin{figure}
    \centering
    \includegraphics[width=5cm]{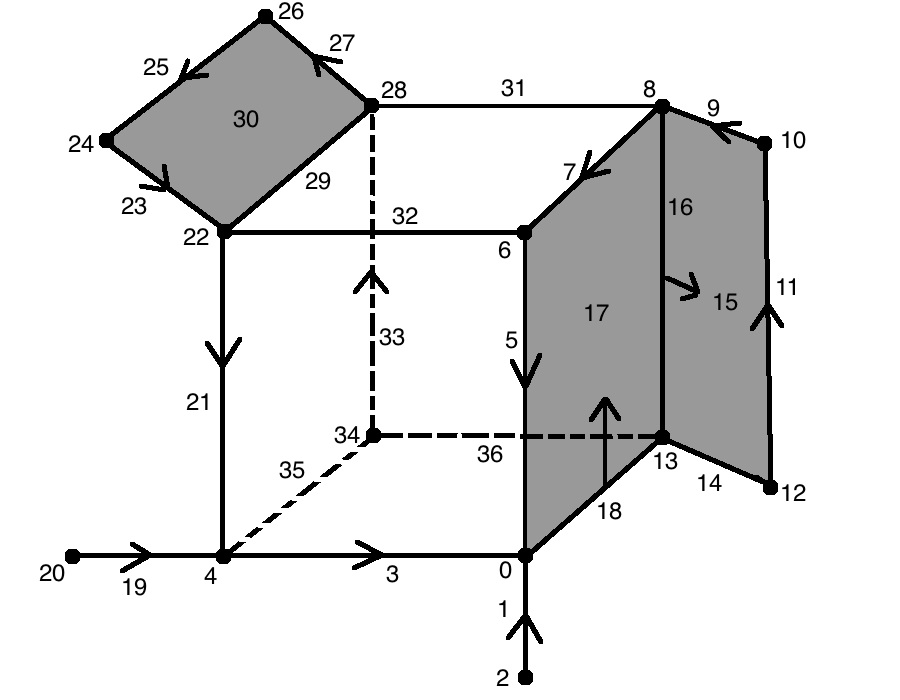}
    \caption{A DVF on a Cubical Complex}
    \label{fig:CubicalDVF}
\end{figure}

The code accompanying \cite{robins2011theory, delgado2015skeletonization} uses the natural DMF on cubical complexes determined by the grayscale values of the vertices in the image, because computers store images as cubical complexes \cite{gonzalez2016encoding, hughes2014computer, kovalevsky1989finite}. Therefore the DVF of interest is also determined by the grayscale values of the vertices of the complex defining a compute graphics image. Figure \ref{fig:CubicalDVF} shows a DVF on a cubical complex with integer values. 

The underlying shape is \emph{connected}.  This means that it is not broken down into several parts, or in other words, there is a path between any two parts of the complex.  Topologically this corresponds to the $0$th Betti number being equal to 1. Moreover, the shape has 5 holes.  This is clear because out of the 8 rectangles, three of them have been filled out as gray, indicating that elementary topological collapses induced by the DVF do not change the topology of the complex. Thus the first Betti number is $5$. Finally, there are no higher dimensional objects in this picture and so all higher Betti numbers are $0$.

Now we show how we can recover the same result by looking a the DMF. The $0$th Betti number corresponds to the critical $0$-cells, which are just the vertices of the cubical complex. Figure \ref{fig:CubicalDVF} has only one critical $0$-cell, namely the one labeled {\it 0}. The first Betti number is determined by critical 1-cells, which we visualize as lines. 

Looking at our DMF we see that the 1-cells with the label set

$$\{ 14,29,31,32,35,36 \}$$ 

are all critical. This does not match our claim that the first Betti number is $5$. However, the rectangle labeled 30 is also critical and has in its boundary the line labeled 29. They thus form a pair that will cancel each other out. Thus we really only have $5$ critical lines and no critical rectangles (equivalently 2-cells). So the DMF confirms that the first Betti number is $5$ and the higher Betti numbers are $0$.  Figure \ref{fig:CubicalDVF}, from a topological viewpoint, is actually a bouquet of 5 circles.

Notice that the DMF also gives us a concrete method to simplify the computations via the DVF. 
The arrows in Figure \ref{fig:CubicalDVF} represent elementary topological collapses induced by the DVF on the cubical complex induced by the values on the vertices. 
By drawing arrows from the higher values to the lower values, we have a concrete method on how to collapse the shape without modifying any crucial information. For example, we can collapse the point labeled {\it 20} onto the point labeled {\it 4} with the path {\it 19} and still have the same topological data.  Clearly we can do some of those simplifications by hand without using a DMF just by looking at the shape.  However, the strength of a DMF is the ability to have a computational method to simplify the structure of a complex in a way that can be implemented algorithmically as in the ANU codebase. Yet we remind the reader that in this manuscript we present methods for moving this functionality beyond grayscale to allow for user-controlled application-specific image analysis, which affects the topological interpretation of an image.

\subsection{Persistent Homology}\label{sec:PH}

{\it Homology groups} are algebraic objects that quantify topological structures of different dimensions present in a cell complex.
The rank of these groups are called {\it Betti numbers}. By comparing the Betti numbers of two cell complexes, we can  determine
how similar the topology of different cell complexes are.
For more details see \cite[2.2]{delgado2015skeletonization}.

In certain situations a cell complex has some additional information, a {\it filtration}.
A filtration of a cell complex $K$ is a chain of cell complexes 
$$ K_0 \subset K_1 \subset K_2 \subset ... \subset K.$$
Intuitively, we think of the filtration as giving us instructions for how to build the actual cell complex $K$. We start with a cell complex $K_0$ and in each step add some piece until we reach the final stage.

For the specific case where the given cell complex has a filtration, we use 
{\it persistent homology} (PH) to compute topological invariants such as the Betti number. Concretely, PH compares the Betti numbers of different filtration 
levels via a {\it barcode diagram} (such as those shown in Figures \ref{fig:Avg_Gray_Kaz_Pst}, \ref{fig:Lum_Gray_Kaz_Pst}, and \ref{fig:Conv_Gray_Kaz_Pst}).  In \cite{cohen2007stability} it is shown that the barcode diagram gives us a accurate description of the Betti numbers of filtered spaces, by proving that a small change in the original space results in a small change in the barcode diagram. Concretely, any given change in the original space gives us a reasonable bound on the change in the barcode diagram.

PH is optimally suited to understand topological features of a cell complex with a given DMF. That is because the integers a DMF assigns to each cell gives the cell complex a filtration. In this case $K_0$ is the the point with the lowest DMF value and each next filtration level adds the lowest dimensional cell which has the next lowest DMF value. For example, in Figure \ref{fig:CubicalDVF} $K_0$ is the point labeled with value 0, $K_1$ consists of the two points with label set $\{ 0,2\}$, and so on. Note we skipped the line labeled $1$ as we have not yet included all lower dimensional cells in our filtration; filtrations are stable under multiple ordering of cells just as the choice of multiple DMFs results in DVFs that reduce a cell complex to the same toplogical information. This filtration allows us to compute the PH of a cell complex. This helps us determine how persistent a Betti number is throughout the filtration. 

In \cite{delgado2015skeletonization} (and in the ANU code) the authors construct a cell complex with a DMF directly computed from the grayscale values in an image and use the mathematical paradigm described above to recover topological features from noisy data and capture key information of the grayscale picture. 

\section{Results and Technical Motivation}\label{sec:results}

As explained in Section \ref{sec:DMT}, DMT allows for the compression of large data sets into critical shapes and points to facilitate the analysis of the homology of the persistent pairs.  Informally, one can think of this compression as a method for identifying the key topological features or ``key shape features" of an object, as discussed in the example of Figure \ref{fig:CubicalDVF}.   In this publication all the objects under study are images stored as RGB arrays that encode images as heat maps representing water scarcity variability (see Section \ref{sec:scarcity}) and crime data variability (see Section \ref{sec:crimedata}).  

To apply the DMT and PH to analyze heat maps, pictures in .jpg form are first converted into grayscale images using three distinct methods for comparison: \emph{Average, Luminosity, and Custom Convertio}, where color values per pixel (where pixels are vertices in a cubical complex) are transformed into grayscale values per pixel, ranging from 0 (black) to 255 (white). The ANU source code from \cite{robins2011theory, delgado2015skeletonization} generates the DMF after assigning 0-cells (pixels) with grayscale values and then choosing integer values for higher-dimesional faces int the cubical complex. Data-informative persistence diagrams are then generated to describe the birth and death time of persistence pairs. With the life-span information of topological features in the input images, patterns of pictures at a specific future time can be predicted and thus the information interpreted from the images can be gained.  The results from our two case analyses follow in Sections \ref{sec:scarcity} and \ref{sec:crimedata}.  Section \ref{sec:scarcity} provides a proof of concept of our methods for enabling user-informed RGB-to-grayscale conversion, while Section \ref{sec:crimedata} provides an proof of concept of how our methods can be used for predictive data analysis.


\subsection{Application I: Analysis of Variability of Water Scarcity}\label{sec:scarcity}
 
\begin{figure}[h!]
    \centering
    \includegraphics[width=5cm, height = 3.2cm]{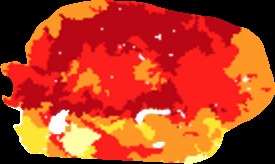}
    \caption{Open-Source Image of Kazakhstan Seasonal Water Risk obtained from the Aqueduct Water Risk Analysis Atlas at http://bit.ly/2fEoU7Q}
    \label{fig:Kaz_original}
\end{figure}

We chose Kazakhstan as a proof of concept for our analysis of the impact of user-defined RGB-to-grayscale conversion because the unusually high variability of temperature and water scarcity in the unique climate of this country leads to topological representations of images with high variability across the images  \cite{yapiyev2017changing}.  Further, the images available at the Aqueduct Water Risk Atlas website from which Figure \ref{fig:Kaz_original} was obtained (See Section \ref{sec:supportingmaterials} for link information) do not even have a uniform RGB weighting on the images, so we analyze the data extracted from the image in Figure \ref{fig:Kaz_original} in detail to make the concepts concrete. 


\begin{figure}[ht!]
    \centering
    \includegraphics[width=5cm, height = 3.2cm]{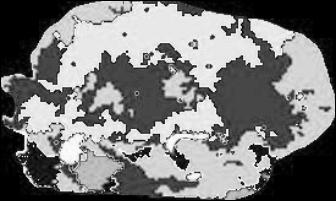}
    \caption{Grayscale Image of Figure \ref{fig:Kaz_original} with Average Method}
    \label{fig:Avg_Gray_Kaz}
\end{figure}

\begin{figure}[ht!]
    \centering
    \includegraphics[width=8.5cm, height = 5cm]{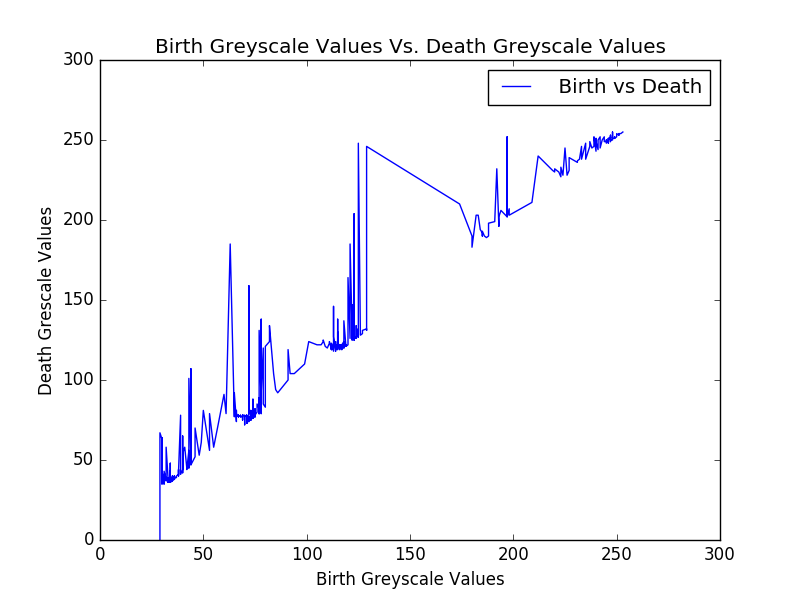}
    \caption{Persistence Diagram with Average Method}
    \label{fig:Avg_Gray_Kaz_Pst}
\end{figure}

Starting with an RGB image of Kazakhstan water scarcity (shown in  (Fig. \ref{fig:Kaz_original}), we used three methods to convert RGB images into Grayscale images.  First, we used an \emph{Average Method}, obtaining the grayscale values by calculating the average of RGB values. Then we obtain the converted grayscale image in Figure \ref{fig:Avg_Gray_Kaz} and the corresponding persistence diagram, shown in Figure \ref{fig:Avg_Gray_Kaz_Pst}.  

\begin{figure}[ht!]
    \centering
    \includegraphics[width=5cm, height = 3.2cm]{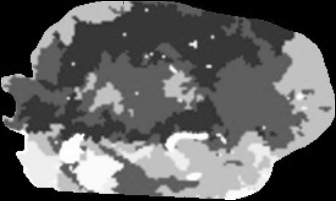}
    \caption{Grayscale Image of \ref{fig:Kaz_original} with Luminosity Method}
    \label{fig:Lum_Gray_Kaz}
\end{figure}

\begin{figure}[ht!]
    \centering
    \includegraphics[width=8.5cm, height = 5cm]{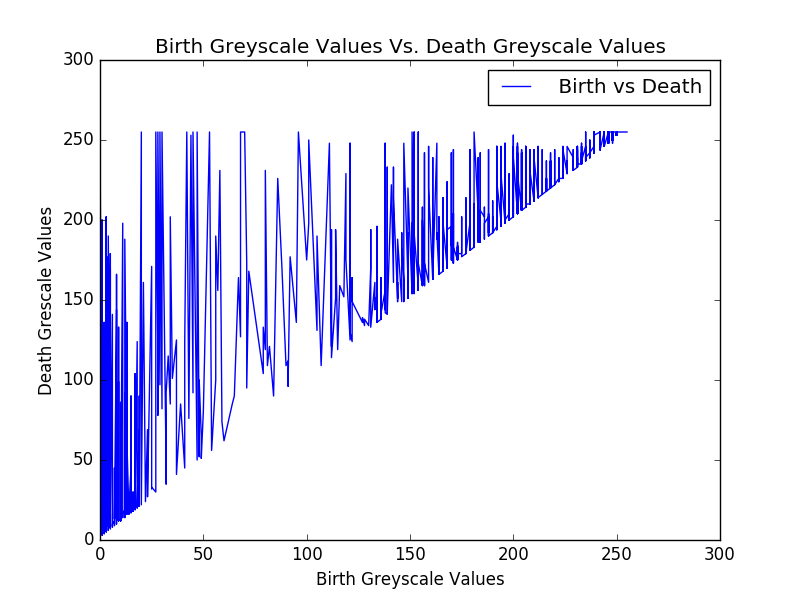}
    \caption{Persistence Diagram with Luminosity Method}
    \label{fig:Lum_Gray_Kaz_Pst}
\end{figure}

Second, we used a \emph{Luminosity Method}, obtaining the grayscale values by assigning different coefficients as \emph{weights} to RGB values respectively to account for human eye's perception. In detail, since human eyes are more sensitive to color green, The value of green is weighted most heavily. The formula for luminosity is $0.21 R + 0.72 G + 0.07 B$. Then we get the converted grayscale image in Figure \ref{fig:Lum_Gray_Kaz} and the corresponding persistence diagram, shown in Figure \ref{fig:Lum_Gray_Kaz_Pst}.

\begin{figure}[ht!]
    \centering
    \includegraphics[width=5cm, height = 3.2cm]{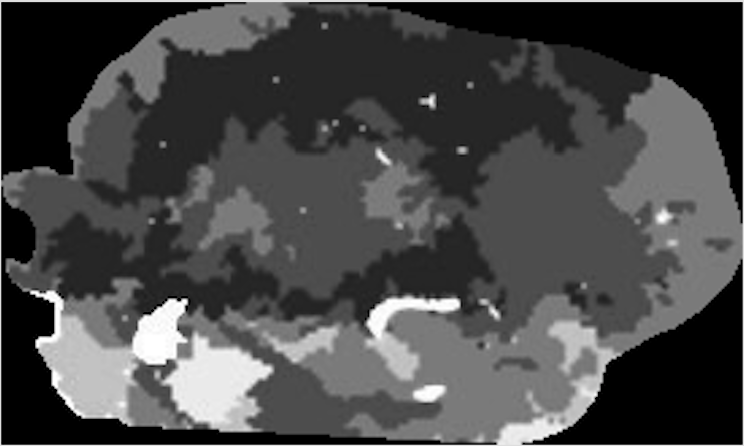}
    \caption{Grayscale Image of Figure \ref{fig:Kaz_original} with Convertio}
    \label{fig:Conv_Gray_Kaz}
\end{figure}

\begin{figure}[ht!]
    \centering
    \includegraphics[width=8.5cm, height = 5cm]{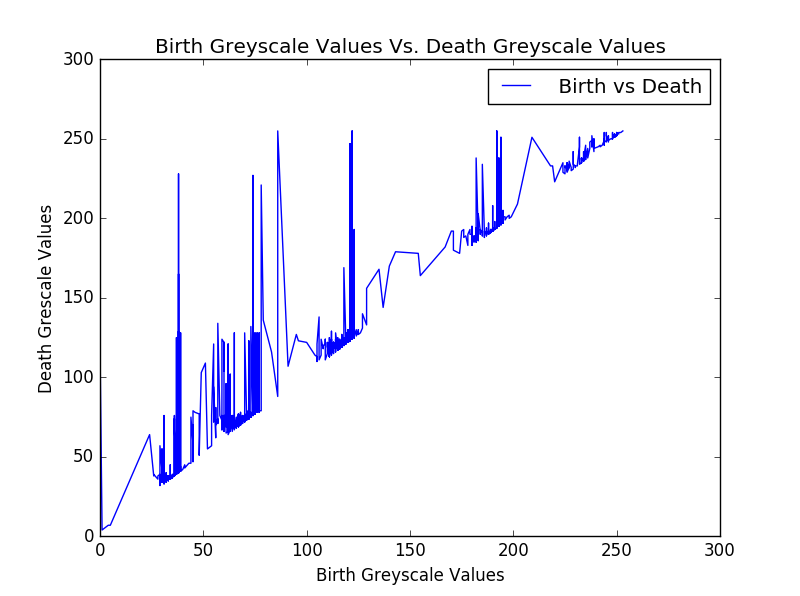}
    \caption{Persistence Diagram with Convertio Method}
    \label{fig:Conv_Gray_Kaz_Pst}
\end{figure}

Third, we used the online converter ``Convertio" (see \hyperref[https://convertio.co/]{https://convertio.co/}) in conjunction with the ANU software. This is the \emph{Custom Convertio} method, which produces a distinct converted grayscale image in Figure \ref{fig:Conv_Gray_Kaz} and corresponding persistence diagram, shown in Figure \ref{fig:Conv_Gray_Kaz_Pst}.

\subsection{Application II: Analysis of Crime Data in Halifax, Nova Scotia}\label{sec:crimedata}


We curated a dataset from Halifax Crime Maps, which is an animated heatmap of density of crimes in Halifax powered by OpenDataHalifax (see \href{Halifax Crime}{http://www.crimeheatmap.ca/}). New data is added weekly leading to transformations in the animation.

\begin{figure}[h!]
    \centering
    \includegraphics[width=7cm, height = 3.2cm]{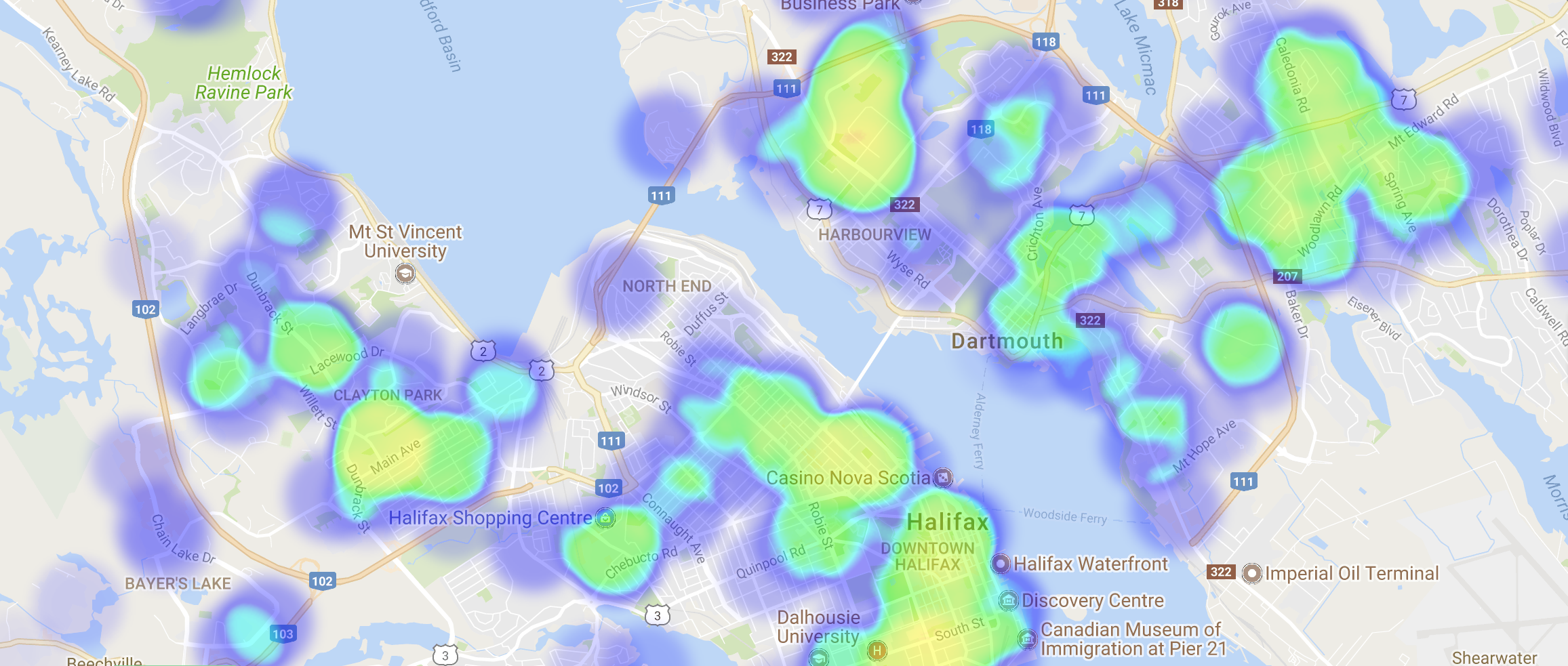}
    \caption{Feb 1st Crime Data (Original)}
    \label{fig:2.1_original}
\end{figure}

To obtain time-sequence data to test predictive analysis, we collected data from OpenDataHalifax dating from February 1st, 2017 to May 24th, 2017, with 15 days between two consecutive images. We cropped the image and retained the part which has the most pattern diversity in the February 1st image, and all the other images are cropped with the same dimensions.  We chose this process to retrieve relevant data for meaningful comparison in a topological sense. 

After cropping the images, we changed the background to black. The crime maps included  information such as local road and river structures that are not relevant to our shape-based data analysis approach. Road and river structures did not change over the time intervals we were studying. Thus we set such background information to black to exclude this data from our analysis and make topologically relevant information easier to extract. 

Next, we changed the color saturation and contrast both to $100\%$ in order to see the significant differences among the RGB weightings, so that diverse patterns could be easily distinguished.  This was a necessary pre-processing step because of the arbitrary nature of RGB weighting in publicly available image data. 

\begin{figure}[ht!]
    \centering
    \includegraphics[width=7cm, height = 3.2cm]{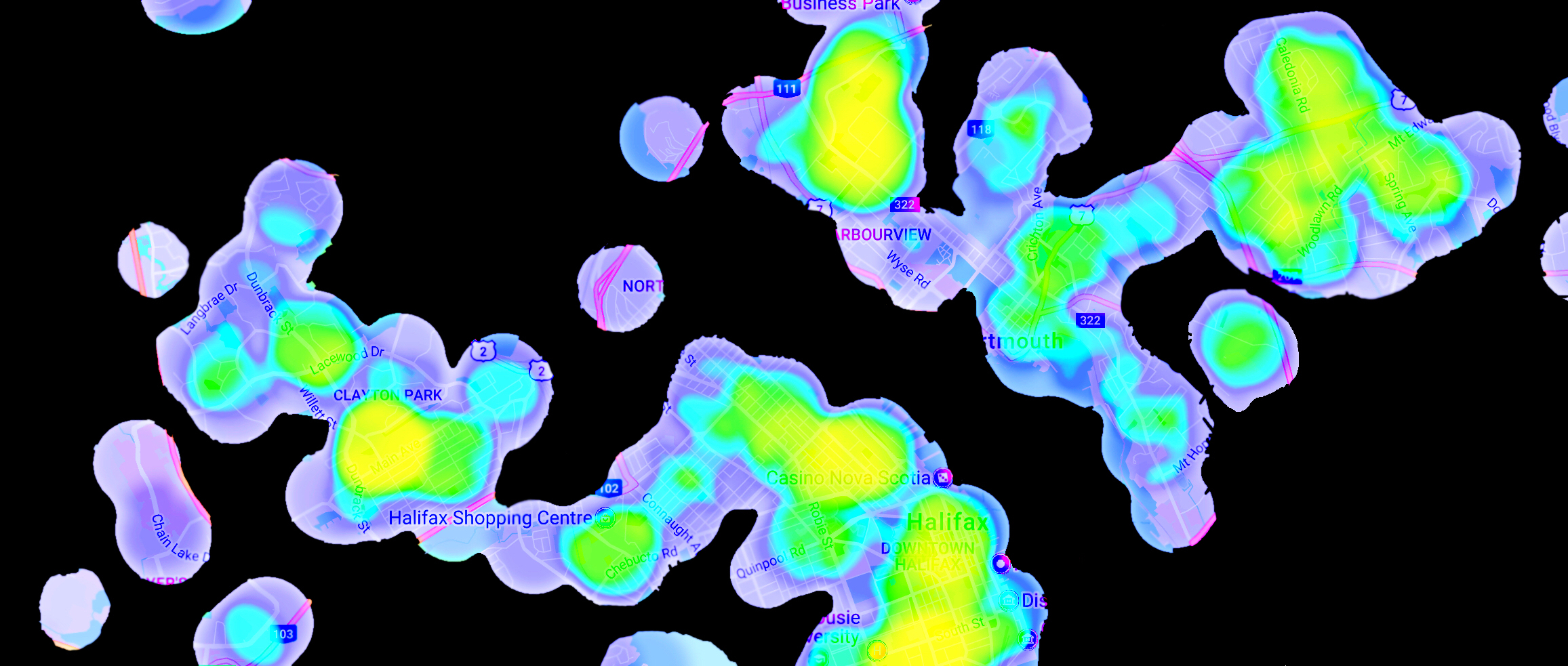}
    \caption{Feb 1st Crime Data with Saturation and Contrast 100\%}
    \label{fig:2.1_100}
\end{figure}

Finally, we transformed all these images from .png files to .pgm files by using the online file type converter “Convertio”, because the ANU code can only use .pgm files as inputs.  See Figures \ref{fig:ANU_flowchart}, \ref{fig:Whole_flowchart}, and  \ref{sec:dataanalysis}.

\begin{figure}[ht!]
    \centering
    \includegraphics[width=7cm, height = 3.2cm]{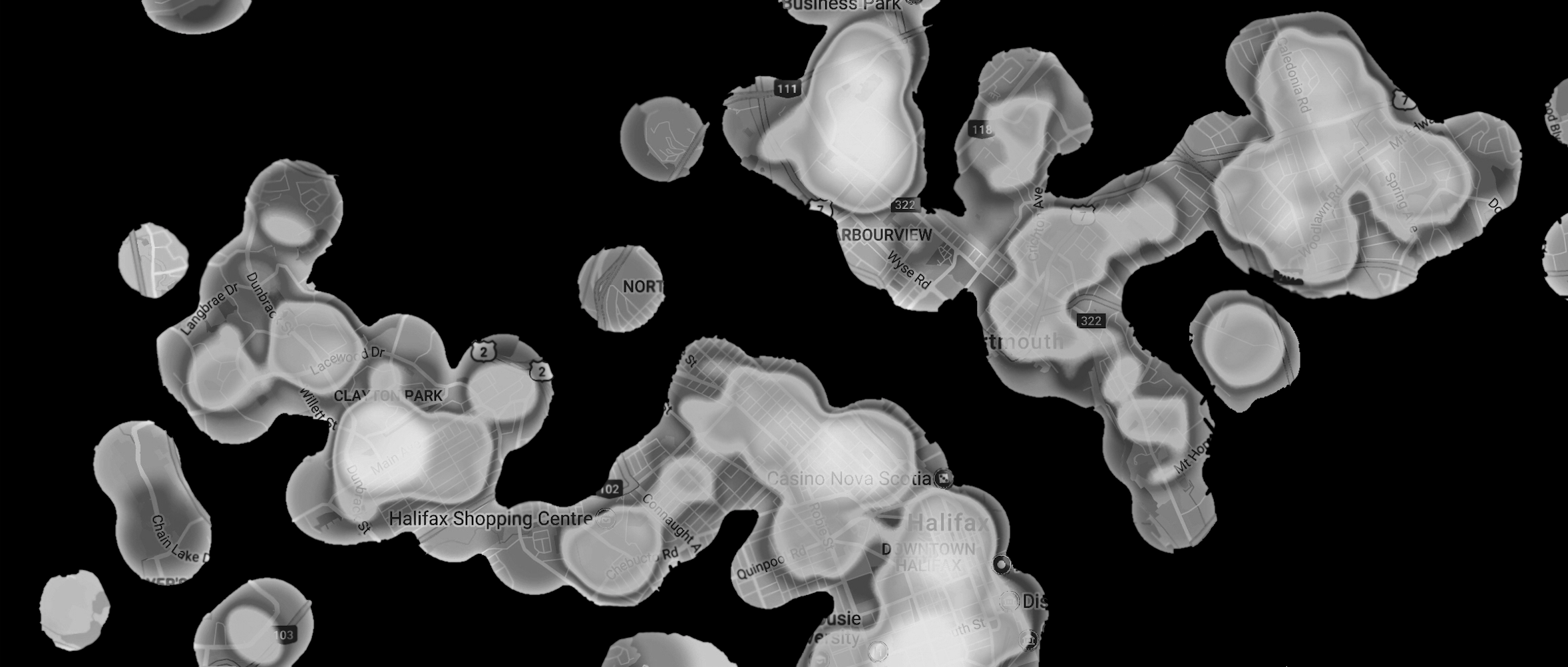}
    \caption{Feb 1st Crime Data in Grayscale}
    \label{fig:2.1_pgm}
\end{figure}

\subsection{Data Processing Procedure Using the ANU Code and Our Code}\label{sec:dataanalysis}

As two codebases are meant to be used in unison to perform similar analyses as those performed in this manuscript, we go into detail regarding the flowcharts in Figures \ref{fig:ANU_flowchart} and \ref{fig:Whole_flowchart} so that the interested user can reproduce our experiments and apply our methods to their own datasets. 

\begin{enumerate}
    \item The .pgm file is first converted to a NetCDF  \hyperref[https://github.com/Unidata/netcdf-c/releases/v4.5.0]{https://github.com/Unidata/netcdf-c/releases/v4.5.0} file via diamorse/util/pgmtonc.C. See \cite{rew1990netcdf} for the original reference. 

\item  Then, the NetCDF file is analyzed for persistence computations via diamorse/python/persistence.py, and outputs a .txt file with entries "birth", "death", "dimension", ``creator $xyz$", ``destructor $xyz$" and ``weight", where \emph{weight} refers to the surjective assignment of an RGB value to a grayscale value. Note that since we are focusing on 1-dimensional homology, the $z$-coordinate simply harmonizes our usage of two distinct codebases.

\item Next, the text file data are sorted and key values are extracted to .csv files for each date of crime data we use correspondingly via TextFiletoCSVFile.java (part of our codebase).

\item The .csv files are imported into Mathematica Notebook (in our codebase) analysis and prediction modeling.

\item In our analysis of the Halifax Crime Data, the Mathematica notebook PredictionModeling.nb read in the .csv files created from the ANU code and pulled out the length of the .csv from May 10th 2017.  In fact, it pulls it out for all of them, but the key variable is the May 10th, 2017, as it is the starting point of the time-series analysis, as well as being a stable starting variable, as outlined in the next point:  

\item PredictionModeling.nb and imports the file from the output .csv files, takes the length of all the .csv files. We chose to do this because the May 10th .csv file has the median length of all six input file sizes. Then we use the starting variable as the May 10th file, but we dropped the May 24th file as it was an outlier in that the number of data points in the .csv file was significantly smaller than the other time-series datasets fof the Halifax crime data. The result of this data analysis decision is shown in Figure \ref{fig:MathematicaPrediction}.  Essentially, it was impossible to recover significant data for our predictive modeling.

\end{enumerate}

\section{Methods and Data Analysis}\label{sec:methodsdata}

The images consumed in our application are available as public image data encoded in RGB values. The ANU code consumes grayscale images, so before using their code, we must pre-process the image to convert the RGB values to grayscale. While there are many common tools available to do this, we found that most conversions used linear combinations of RGB values that resulted in a loss of data regarding the key shape or topological information of the cell complex (see \cite{kanan2012color} for a review of unresolved issues identified with this problem). 

The consideration of these linear combinations is important as the weighting of RGB values affects the result of the grayscale conversion in a manner that directly influences the homological features of the converted images. In particular, the use of many linear-combination-based RGB tools in our early experiments (evidenced best by the Average Method in Section \ref{sec:scarcity}) led to over-weighting of dark grayscale values resulting from the use of such conversion tools erased the birth and death of homological features of RGB data that were essential to a informative topological analysis. Figure \ref{fig:Avg_Gray_Kaz_Pst} shows the result of the over-weighting of dark grayscale values both in the longer lifetimes of trivial features in Figure \ref{fig:Avg_Gray_Kaz} as well as the brief lifetime of of the first grayscale value at about 25 on the $x$-axis.  

In particular, too much information is lost about the 1-dimensional homology of an image under analysis if a na\"{i}ve RGB-to-grayscale conversion is used. The Luminosity Method for this conversion as displayed in the diagram in Figure \ref{fig:Lum_Gray_Kaz_Pst} preserves the most information about birth and death of 1-dimensional homology out of the three methods tested on the Kazakhstan water scarcity data. We argue that shorter birth and death cycles for 1-dimensional homology tells the user of our methods more information usable for custom data analysis. 

Comparing Figures \ref{fig:Avg_Gray_Kaz_Pst} and \ref{fig:Conv_Gray_Kaz_Pst}, one sees that the Custom Covertio Method does not lose as much life-cycle information for topological features as the Average Method. The Luminosity Method picks up more birth-death cycling than the Custom Covertio method, which in turn picks up more birth-death cycling than the Average Method. Therefore we argue that the Luminosity Method is best suited to the analysis of this image.

\subsection{Mathematica Notebook Calculations and Predictive Analysis}

The Mathematica notebook PredictionModeling.nb available on our github (See Section \ref{sec:supportingmaterials}) reads through all of the dates encoded in the .csv files we extracted from the Halifax Crime Data website and randomly samples which values to take out of the datasets, as they are too large to enable a full sampling with our computational resources. The number of data points is fixed by the number of data points from the May 10th sample. We chose May 10th as the basepoint proof of concept for this analysis as it had the second fewest number of points.  When the notebook samples the data from the .csv files it pulls out the grayscale values from when the pattern was destroyed and created. 

The prediction analysis relies on six date-based values explained in Section \ref{sec:crimedata}.  The notebook consumes the six date-based values after the three RBG-to-grayscale transformations are completed- i.e. the six places where significant features were born or destroyed. 
The choice of six dates were due to the limitations of our computational resources.  

The code in PredictionModeling.nb took six values for six dates and plotted them onto an x and y plot, as visualized in Figure \ref{fig:MathematicaPrediction}, and looks for significant data points on the plot. This is computationally expensive as the average computing time was on the order of 10 hours for each data point on an Ubuntu Linux operating system (Version 16).  

The PredictionModeling.nb  notebook creates a list of 35000 functions and then with functions with respect to the $x$ value and the $y$ value as a function of $x$. There is no $z$-value in our applications because we are  only looking at 1-dimensional homology in the images we analyze. In PredictionModeling.nb the $x$ axis represents the number of data point collected, and the $y$ axis represents how long the pattern persisted.

For predictive analysis, PredictionModeling.nb first imports the nine .csv files for the nine dates as nine separate table variables. The notebook then finds the length for eight of the nine tables. The May 24th Table (an input data point) was dropped due to how small it was.  (See the zero value on the plot in Figure \ref{fig:MathematicaPrediction}).  PredictionModeling.nb then stores the May 10th length in a separate variable called "keyLength." May 10th is the \emph{key length} because its table was the smallest after the May 24th table was dropped. 

The next process in PredictionModeling.nb is to randomly sample from each of the eight dates by sampling their index values. The number of values sampled from is equal to the integer value of the "keyLength." The value stored in keyLength is about 35,000. The index that is pulled out is used to figure out where in the table should the birth and death grayscale values be extracted from. These values are then placed into eight more tables in Mathematica. The death grayscale value is then subtracted from the birth grayscale value. This new value is stored into eight more tables, one for each of the dates where information was collected.

These eight tables are then compiled into one table: a table of tables of integer values.  Note, the number eight has nothing to do with the number of data points collected and is purely a computational issue. Next, the value from the first index in each of the tables for the first six dates is extracted and placed into a separate set. This is done until the keyLength value is met and there are no more points to analyze. 

The sets are then plotted with the $x$-axis being the date value and the $y$-axis being the grayscale value for the amount of time that the pattern persisted. It does this once for each pattern because it is unnecessary to use create graphics at each step within the Mathematica computations. Graphics such as Figures \ref{fig:ANU_flowchart}, \ref{fig:Whole_flowchart} and \ref{fig:MathematicaPrediction} are produced solely to understand code functionality.

After the plot such as Figure \ref{fig:MathematicaPrediction} has been created, the PredictionModeling.nb attempts to find a predictive function using only a single independent variable for the 35,000 sets. The independent variable, $x$, is extracted from the data for a date we collected: e.g. (1 = Feb 01, 2 = Feb 15, 3 = Mar 01... 6 = Apr 12) and the dependent variable, $y$, is the predicted length of the lifespan of the pattern for that date for the comparison of index 1. 

After the 35,000 functions have been recovered, they are saved to a table. The first function is pulled out of the table and the $x$ variable is replaced with 7 because that is the next value to come; logically this would be at a 15-day interval according to our data curation methods as outlined in Section \ref{sec:crimedata}.  We used six dates to train the predictive function and the seventh date is a predictive test value. PredictionModeling.nb solves for what the $y$-value (the lifespan of the pattern for that date for the comparison of index 1) of the seventh date is expected to be and then compares that to the actual seventh date that we have on record. PredictionModeling.nb looks for a percent error. 
 
As we are not aware of other tools for predictive RGB data prediction methods, it is up to the consumer of this publication to decide a tolerable error percentage. Yet a reasonable recommendation would be to say if the error was was less than five percent, this would be a success: this is a high bound as PredictionModeling.nb would run this computation for each of the 35,000 sets of functions generated by the notebook. There is no reason besides computational resource access at our institution that more than a 7th date could be predicted.

\subsection{Using Our Methods with the ANU Codebase}\label{sec:usandANU}

\begin{figure}
    \centering
    \includegraphics[width=8.5cm]{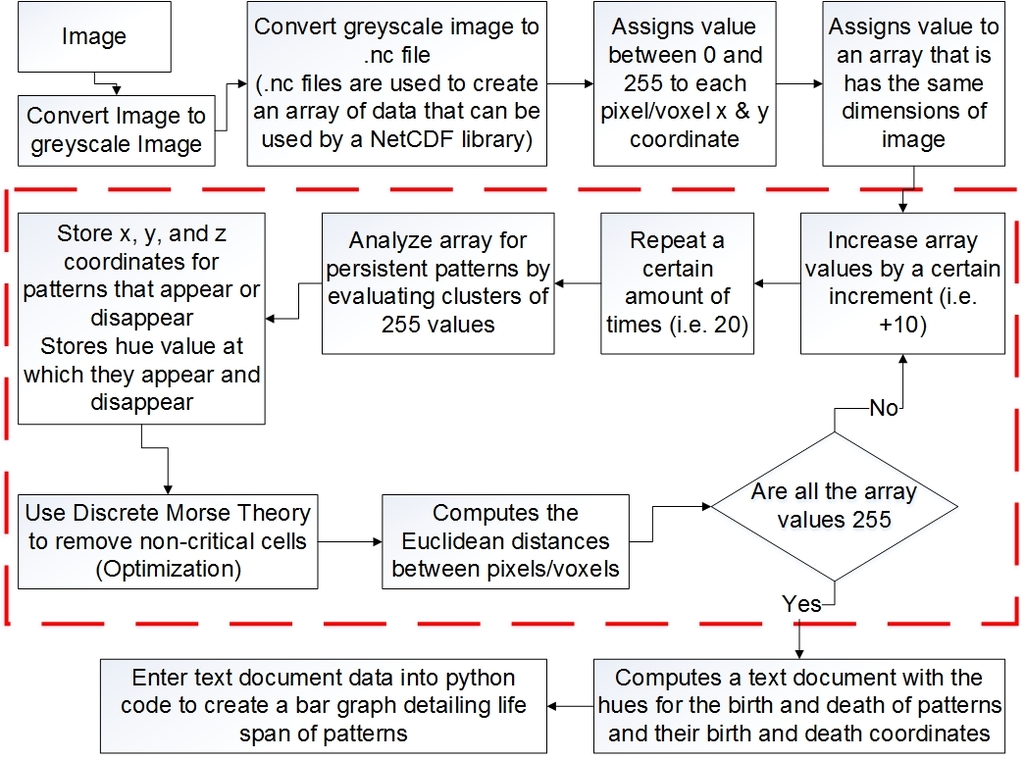}
    \caption{Flowchart for Data Analysis Using ANU Codebase and Our Codebase}
    \label{fig:ANU_flowchart}
\end{figure}

\begin{figure}
    \centering
    \includegraphics[width=8.5cm]{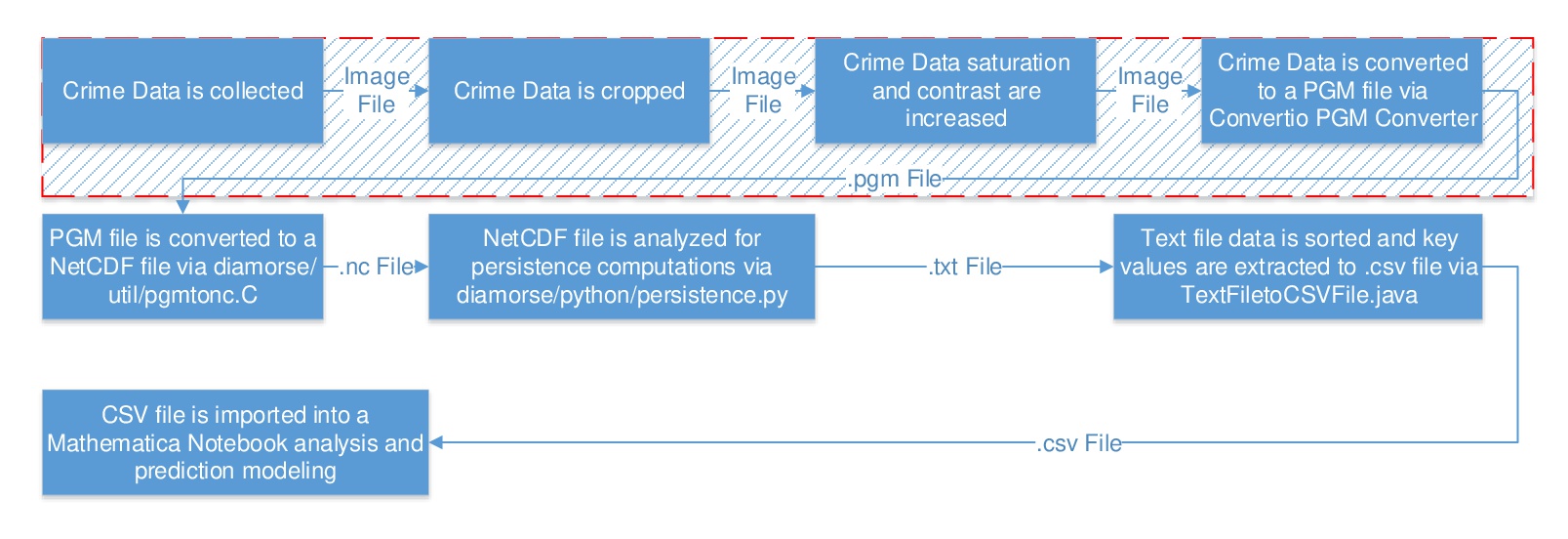}
    \caption{Flowchart for How ANU Code and our Codebase Work Together for Predictive Analysis for Crime Data}
    \label{fig:Whole_flowchart}
\end{figure}

In Figure \ref{fig:ANU_flowchart} we provide a visual representation of how to combine our code with the ANU code for the methods in Section \ref{sec:scarcity}. Our codebase handles the conversion of an RGB image to a grayscale image (pictured in the upper left-hand corner of the Figure) so that we can employ the functionality of the ANU code. In Figure \ref{fig:Whole_flowchart} it shown how to use the two codebases in conjunction regarding the methods explained in Section \ref{sec:crimedata}. 

The ANU code takes in an image in the .pgm format and then it converts that image to an image in the NETCDF3 format. The code functions in two distinct ways. The first option is that it creates a segmented black and white image and sets the pixels (one-dimensional components of the cubical complex in question) with non-zero values to one. The second option is that the code sets the values of certain pixels, whose values are above a certain specified value, to one.

After doing this the code then analyzes the image in the NETCDF3 format to generate data identifying the persistent pairs in the cubical complex. This is accomplished by running the picture through the persistence.py script available in the ANU codebase. This process is outlined in Figure \ref{fig:ANU_flowchart}.

The ANU code then writes the data in a .txt file. Using our code (See Figure \ref{fig:Whole_flowchart}, the .txt file is then inserted into a Java converter that changes the .txt file to a .csv file. The .csv file is then given as input to our notebook PredictionModeling.nb, which then takes random points from the dataset and generates functions to predict future trends.  

\subsection{Key Commands Used in Data Analysis from the ANU Codebase}\label{sec:commands}




We explain here in detail the Custom Convertio Method as outlined in Section \ref{sec:scarcity}. The image.pgm file is an image file that was converted to a .pgm file through Convertio. In the predictive application of this research project, the images are taken in the form of .jpeg files from screenshots of the Hailfax Crime Data. Explanation of how these images are processed before being used in the ANU code can be found in Section \ref{sec:crimedata}.

\begin{verbatim}
Input File: image.pgm
Output File: tomo_floatimage.nc
Command: diamorse ./bin/pgmtonc image.pgm
\end{verbatim}

The command pgmtonc uses the C source code from the ANU codebase to convert the image.pgm file to the tomofloatimage.nc. The tomofloatimage.nc file is the image file in the NetCDF form. It contains the information present in the image.pgm file in a multidimensional array of integer values. This tomofloatimage.nc file is used in the persistence.py code to create the text file with information on the persistence pairs in the image.pgm file.

We ran the pgmtonc command using terminal on a Linux Ubuntu laptop (Version 14.04) after we changed the directory to the folder containing the folder "bin" with the pgmtonc.c code inside of it. The image.pgm file was in the same directory as the directory with the "bin" folder. This directory was called "diamorse" for us.  Filepaths and directory names will of course need to be adjusted depending on the user of our software. 

This command was ran for each of the six images pulled in from HailfaxCrimeData because we needed to have separate tomofloatimage.nc files for each of them. To identify which files were which, each .pgm file name was given the month and date of when the data was taken. This file naming automatically came over when the .nc file was created.




\begin{verbatim}
Input File: tomo_floatimage.nc
Output File: file.txt
Command: diamorse ./python/persistence.py 
-t 1.0 -r tomo_floatimage.nc > file.txt
\end{verbatim}


The text file that is output from the persistence.py source code gives the information on the persistent pairing in a tabular form. This information is the the grayscale values for when the persistent pair is created and destroyed, the persistent pair's dimension, and the $xyz$-coordinate locations for when the persistent pair is created and destroyed. The grayscale values for the birth and death of the persistent pairs is what is used in the scope for this project.

We ran the persistence.py command using terminal on Linux Ubuntu Version 14.04. The tomofloatimage.nc file from the above documentation, ./bin/pgmtonc, should be output to the "diamorse" folder and does not need to be moved. When the command is run, the target location is maintained as the "diamorse" folder and the target file type was a text file. We ran this command nine times for each of the tomofloatimage.nc files created from the previous command. The name of the target file was changed to the month and date to match the initial image name.

The grayscale values contained in the text files are the same information that can be used to create figures similar to Figure 4, Figure 6, and Figure 8. In those figures, the vertical bars were created by using the grayscale value for the birth of the persistent pair as the bottom point of the vertical bar and using the grayscale value for the death of the persistent pair as the top of the bar. Sections of the diagram that were not vertical when the birth and death grayscale values were the same. This is apparent in Figure 4 between the 140 and 170 on the birth grayscale Values.

\subsection{Key Commands Used in Data Analysis from our Codebase}\label{sec:commandsused}

 
The text file from the ANU's persistence.py source code is the input for the TextFiletoCSVFile.java. The java code was written so that it would read in all of the nine text documents that we created at the same time and create nine separate .csv files with the name from the text documents.

We ran the java source code on a Windows 10 machine using the Eclipse IDE with the Java Platform, Standard Edition Development Kit version 9.

\begin{verbatim}
Input File: file.txt
Output File: file.csv
Command: ./TextFiletoCSVFile.java
\end{verbatim}


The file.csv file contains the grayscale value for when the when the persistent pair is created and destroyed from the text file generated from the ANU codebase and it indexes each of the persistent pairs. The index starts at one and finishes at the last entry from the text file. The .csv file was designed to have the first column as the index number, the next column had the grayscale value that the persistent pair was created, and the third and last column had the grayscale value that the persistent pair was destroyed.

Creating .csv files from the text files was done because the information could be easily read into Mathematica and would make the Mathematica notebook easier to write and follow.  

\begin{verbatim}
Input File: file.csv
Output File: notebook.nb
\end{verbatim}


PredictionModeling.nb contains a plot of the first pattern over the first six dates, a list of functions for each of the data sets, an estimate for the lifelines to happen in the next date, and a percent error between the estimate and the actual grayscale values. 

\begin{figure}
    \centering
    \includegraphics[width=8.5cm]{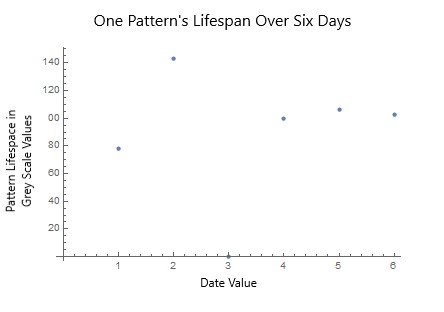}
    \caption{Mathematica Prediction Diagram}
    \label{fig:MathematicaPrediction}
\end{figure}

After point six in Figure \ref{fig:MathematicaPrediction} point 7 should indicate as a function of $x$ in the $y$-coordinate. We solve the equation, and give that position in the $y$-coordinate as a function of $x$ only. Note that in Figure \ref{fig:MathematicaPrediction} the date 3 is valued zero because the image had no recoverable topological patterns on that date.

\section{Discussion and Future Directions}\label{sec:discussion}

The goal of this manuscript is to contribute to a path towards predictive topological data analysis based only on RGB images (either publicly or privately available) to improve computational feasibility in an age of an overabundance of visual data. The contribution of this manuscript is the development of methods to bypass computational techniques including parallelized machine learning \cite{youkana2017parallelization, reina2016effective}, develop new methods for uncovering statistical properties of image-encoded data \cite{robins2016percolating}, and add potential for application-based directions for the theory in algebraic publications such as \cite{allili2017reducing}.   

\subsection{Relevance and Applications of Data Analysis in this Manuscript}\label{sec:datadiscussion}

The proof of concept of our methods established in Section \ref{sec:scarcity} is that RGB-to-grayscale conversions affect the amount of toplogical features recovered from an image. If time-series data was available for the Kazakhstan water-scarcity image data it is reasonable to assume that water scarcity in this volatile region would be predictable in the near term. 

The proof of concept for predictive image-based analysis was established in Section \ref{sec:crimedata} because time-series data was accessible via open access collection by the authors for Halifax, Nova Scotia crime patterns over regular time intervals.


\subsection{Future Questions and Directions}\label{sec:future}

\begin{enumerate}
    \item In the future we expect that image-based DMT data analysis could be applied to water scarcity or other climactic conditions. This could be promising for any open data source updated at more regular intervals. In an era of rapidly fluctuating climate, additional predictive tools that require minimal computational time should be useful. 
    \item To allow for concrete data analysis using our methods, we edited the pictures in Section \ref{sec:scarcity}.  In particular, we encountered issues with variation in the dimension that we determined were a byproduct of the non-informative background of the time-series crime data. Due to our editing of the the original images from the crime data, we may have introduced bias in the predictive methods, though this is unlikely, as there was no data in the background color, only user-interface based RGB graphics design. Yet for other types of data this question should be investigated. 
    \item Yet for other images it may be the case that a custom surjection from the RGB scale of $\{0 \ldots 255\}^{3}$ onto the grayscale array of $\{0 \ldots 255 \}$ may be necessary.  Hence the inclusion of the script SurjectionCode.py in our code release (See Section \ref{sec:supportingmaterials}) to allow for the application-informed user to customize a mapping (a surjection) onto grayscale values from RGB values that extracts informative birth-death patterns in a barcode diagram.  This script is still under development. The goal of this script is to provide a customized user-informed RGB weighting to use with the rest of our codebase, which is still under active development. 
    \item Running PredictionModeling.nb on a laptop using Ubuntu Verson 14.04, comptuations are not significantly refined further after taking 10, 15, or 50 points and the the PredictionModeling.nb notebook stalls before 9 hours. Errors in our computations showed that limitations of our server access at our home university prevented a complete analysis.  With better computational resources we would be able to execute the computational goals of our theoretical development. It is our hope that users with large-scale computing resources (i.e. campus computing clusters instead of desktop or laptop machines) will be able to perform more detailed analyses with our codebase. 

\end{enumerate}

\section{Supporting Materials and Software}\label{sec:supportingmaterials}

The data used in Section \ref{sec:scarcity} can be found at Aqueduct Water Risk Atlas website: 

\href{The Aqueduct Water Risk Atlas website}{http://www.wri.org/applications/maps/aqueduct-atlas/}

The data used in Section \ref{sec:crimedata} can be found at Halifax Crime's website: 
\href{Halifax Crime}{http://www.crimeheatmap.ca/}

All novel code developed for this manuscript and used in our data analysis can be found at the github repository \hyperref[https://github.com/redavids/IBTCDA/tree/master]{https://github.com/redavids/IBTCDA/tree/master}.  This code requires dependencies explained on the the github repository, including the repository released with the publications \cite{robins2011theory, delgado2015skeletonization}. Installation instructions are available at the github.

\section*{Acknowledgments} This project was supported by a Mathways Grant NSF DMS-1449269 to the Illinois Geometry Lab. We thank Tyler Graham and Titan Wibowo for their participation and comments during early phases of the project. R.D. was supported by NSF grant DMS-1401591.  

\ifCLASSOPTIONcaptionsoff
 \newpage
\fi

\bibliographystyle{IEEEtran}
\bibliography{refs.bib}

\begin{thebibliography}{10}
\providecommand{\url}[1]{#1}
\csname url@samestyle\endcsname
\providecommand{\newblock}{\relax}
\providecommand{\bibinfo}[2]{#2}
\providecommand{\BIBentrySTDinterwordspacing}{\spaceskip=0pt\relax}
\providecommand{\BIBentryALTinterwordstretchfactor}{4}
\providecommand{\BIBentryALTinterwordspacing}{\spaceskip=\fontdimen2\font plus
\BIBentryALTinterwordstretchfactor\fontdimen3\font minus
  \fontdimen4\font\relax}
\providecommand{\BIBforeignlanguage}[2]{{%
\expandafter\ifx\csname l@#1\endcsname\relax
\typeout{** WARNING: IEEEtran.bst: No hyphenation pattern has been}%
\typeout{** loaded for the language `#1'. Using the pattern for}%
\typeout{** the default language instead.}%
\else
\language=\csname l@#1\endcsname
\fi
#2}}
\providecommand{\BIBdecl}{\relax}
\BIBdecl

\bibitem{edelsbrunner2008persistent}
H.~Edelsbrunner and J.~Harer, ``Persistent homology-a survey,''
  \emph{Contemporary {M}athematics}, vol. 453, pp. 257--282, 2008.

\bibitem{reininghaus2015stable}
J.~Reininghaus, S.~Huber, U.~Bauer, and R.~Kwitt, ``A stable multi-scale kernel
  for topological machine learning,'' in \emph{Proceedings of the {IEEE}
  {C}onference on {C}omputer {V}ision and {P}attern {R}ecognition}, 2015, pp.
  4741--4748.

\bibitem{robins2016percolating}
V.~Robins, M.~Saadatfar, O.~Delgado-Friedrichs, and A.~P. Sheppard,
  ``Percolating length scales from topological persistence analysis of
  micro-{CT} images of porous materials,'' \emph{Water {R}esources {R}esearch},
  vol.~52, no.~1, pp. 315--329, 2016.

\bibitem{lee2012persistent}
H.~Lee, H.~Kang, M.~K. Chung, B.-N. Kim, and D.~S. Lee, ``Persistent brain
  network homology from the perspective of dendrogram,'' \emph{{IEEE}
  transactions on medical imaging}, vol.~31, no.~12, pp. 2267--2277, 2012.

\bibitem{bendich2016persistent}
P.~Bendich, J.~Marron, E.~Miller, A.~Pieloch, and S.~Skwerer, ``Persistent
  homology analysis of brain artery trees,'' \emph{The annals of applied
  statistics}, vol.~10, no.~1, p. 198, 2016.

\bibitem{gamble2010exploring}
J.~Gamble and G.~Heo, ``Exploring uses of persistent homology for statistical
  analysis of landmark-based shape data,'' \emph{Journal of {M}ultivariate
  {A}nalysis}, vol. 101, no.~9, pp. 2184--2199, 2010.

\bibitem{chung2009persistence}
M.~K. Chung, P.~Bubenik, and P.~T. Kim, ``Persistence diagrams of cortical
  surface data,'' in \emph{International {C}onference on {I}nformation
  {P}rocessing in {M}edical {I}maging}.\hskip 1em plus 0.5em minus 0.4em\relax
  Springer, 2009, pp. 386--397.

\bibitem{mischaikow2013morse}
K.~Mischaikow and V.~Nanda, ``Morse theory for filtrations and efficient
  computation of persistent homology,'' \emph{Discrete \& {C}omputational
  {G}eometry}, vol.~50, no.~2, pp. 330--353, 2013.

\bibitem{cerri2013betti}
A.~Cerri, B.~D. Fabio, M.~Ferri, P.~Frosini, and C.~Landi, ``Betti numbers in
  multidimensional persistent homology are stable functions,''
  \emph{Mathematical {M}ethods in the {A}pplied {S}ciences}, vol.~36, no.~12,
  pp. 1543--1557, 2013.

\bibitem{robins2011theory}
V.~{R}obins, P.~J. {W}ood, and A.~P. {S}heppard, ``Theory and algorithms for
  constructing discrete {M}orse complexes from grayscale digital images,''
  \emph{{IEEE} {T}ransactions on pattern analysis and machine intelligence},
  vol.~33, no.~8, pp. 1646--1658, 2011.

\bibitem{delgado2015skeletonization}
O.~{D}elgado {F}riedrichs, V.~{R}obins, and A.~{S}heppard, ``Skeletonization
  and partitioning of digital images using discrete {M}orse theory,''
  \emph{{IEEE} transactions on pattern analysis and machine intelligence},
  vol.~37, no.~3, pp. 654--666, 2015.

\bibitem{priestley1972assessment}
C.~Priestley and R.~Taylor, ``On the assessment of surface heat flux and
  evaporation using large-scale parameters,'' \emph{Monthly weather review},
  vol. 100, no.~2, pp. 81--92, 1972.

\bibitem{xingjian2015convolutional}
S.~Xingjian, Z.~Chen, H.~Wang, D.-Y. Yeung, W.-K. Wong, and W.-c. Woo,
  ``Convolutional {LSTM} network: A machine learning approach for precipitation
  nowcasting,'' in \emph{Advances in neural information processing systems},
  2015, pp. 802--810.

\bibitem{gill2014bayesian}
J.~Gill, \emph{Bayesian methods: {A} social and behavioral sciences
  approach}.\hskip 1em plus 0.5em minus 0.4em\relax {CRC} press, 2014, vol.~20.

\bibitem{oki2006global}
T.~Oki and S.~Kanae, ``Global hydrological cycles and world water resources,''
  \emph{{S}cience}, vol. 313, no. 5790, pp. 1068--1072, 2006.

\bibitem{smakhtin2004pilot}
V.~Smakhtin, C.~Revenga, and P.~D{\"o}ll, ``A pilot global assessment of
  environmental water requirements and scarcity,'' \emph{Water
  {I}nternational}, vol.~29, no.~3, pp. 307--317, 2004.

\bibitem{petit2016paradise}
O.~Petit, ``Paradise lost? {T}he difficulties in defining and monitoring
  {I}ntegrated {W}ater {R}esources {M}anagement indicators,'' \emph{Current
  opinion in environmental sustainability}, vol.~21, pp. 58--64, 2016.

\bibitem{nakaya2010visualising}
T.~Nakaya and K.~Yano, ``Visualising {C}rime {C}lusters in a {S}pace-time
  {C}ube: {A}n {E}xploratory {D}ata-analysis {A}pproach {U}sing {S}pace-time
  {K}ernel {D}ensity {E}stimation and {S}can {S}tatistics,'' \emph{Transactions
  in {GIS}}, vol.~14, no.~3, pp. 223--239, 2010.

\bibitem{de2017automatic}
M.~H. de~Boer, H.~Bouma, M.~C. Kruithof, F.~B. ter Haar, N.~M. Fischer, L.~K.
  Hagendoorn, B.~Joosten, and S.~Raaijmakers, ``Automatic analysis of online
  image data for law enforcement agencies by concept detection and instance
  search,'' in \emph{Counterterrorism, {C}rime {F}ighting, {F}orensics, and
  {S}urveillance {T}echnologies}, vol. 10441.\hskip 1em plus 0.5em minus
  0.4em\relax International {S}ociety for {O}ptics and {P}hotonics, 2017, p.
  104410H.

\bibitem{forman2002user}
R.~Forman, ``A user’s guide to discrete {M}orse theory,'' \emph{S{\'e}m.
  {L}othar. {C}ombin}, vol.~48, p. 35pp, 2002.

\bibitem{milnor1959spaces}
J.~Milnor, ``On spaces having the homotopy type of a {CW}-complex,''
  \emph{Transactions of the {A}merican {M}athematical {S}ociety}, vol.~90,
  no.~2, pp. 272--280, 1959.

\bibitem{gonzalez2016encoding}
R.~Gonzalez-Diaz, M.-J. Jimenez, and B.~Medrano, ``Encoding specific 3{D}
  polyhedral complexes using 3{D} binary images,'' in \emph{International
  {C}onference on {D}iscrete {G}eometry for {C}omputer {I}magery}.\hskip 1em
  plus 0.5em minus 0.4em\relax Springer, 2016, pp. 268--281.

\bibitem{hughes2014computer}
J.~F. Hughes, \emph{Computer graphics: principles and practice}.\hskip 1em plus
  0.5em minus 0.4em\relax Pearson {E}ducation, 2014.

\bibitem{kovalevsky1989finite}
V.~A. Kovalevsky, ``Finite topology as applied to image analysis,''
  \emph{Computer vision, graphics, and image processing}, vol.~46, no.~2, pp.
  141--161, 1989.

\bibitem{cohen2007stability}
D.~Cohen-Steiner, H.~Edelsbrunner, and J.~Harer, ``Stability of persistence
  diagrams,'' \emph{Discrete \& {C}omputational {G}eometry}, vol.~37, no.~1,
  pp. 103--120, 2007.

\bibitem{yapiyev2017changing}
V.~Yapiyev, Z.~Sagintayev, A.~Verhoef, A.~Kassymbekova, M.~Baigaliyeva,
  D.~Zhumabayev, D.~Malgazhdar, D.~Abudanash, N.~Ongdas, and S.~Jumassultanova,
  ``The changing water cycle: {B}urabay {N}ational {N}ature {P}ark, {N}orthern
  {K}azakhstan,'' \emph{Wiley {I}nterdisciplinary {R}eviews: {W}ater}, 2017.

\bibitem{rew1990netcdf}
R.~Rew and G.~Davis, ``{N}et{CDF}: an interface for scientific data access,''
  \emph{{IEEE} computer graphics and applications}, vol.~10, no.~4, pp. 76--82,
  1990.

\bibitem{kanan2012color}
C.~Kanan and G.~W. Cottrell, ``Color-to-grayscale: does the method matter in
  image recognition?'' \emph{Plo{S} one}, vol.~7, no.~1, p. e29740, 2012.

\bibitem{youkana2017parallelization}
I.~Youkana, J.~Cousty, R.~Saouli, and M.~Akil, ``Parallelization strategy for
  elementary morphological operators on graphs: distance-based algorithms and
  implementation on multicore shared-memory architecture,'' \emph{Journal of
  {M}athematical {I}maging and {V}ision}, pp. 1--25, 2017.

\bibitem{reina2016effective}
R.~Reina-Molina, D.~D{\'\i}az-Pernil, P.~Real, and A.~Berciano, ``Effective
  homology of k-{D} digital objects (partially) calculated in parallel,''
  \emph{Pattern {R}ecognition {L}etters}, vol.~83, pp. 59--66, 2016.

\bibitem{allili2017reducing}
M.~Allili, T.~Kaczynski, and C.~Landi, ``Reducing complexes in multidimensional
  persistent homology theory,'' \emph{Journal of {S}ymbolic {C}omputation},
  vol.~78, pp. 61--75, 2017.

\end{thebibliography}

\end{document}